\documentclass[preprint,12pt]{elsarticle}

\usepackage{graphicx}
\usepackage{caption}
\usepackage{subcaption}
\usepackage{amsmath}
\usepackage{color,soul} 
\usepackage[table]{xcolor}
\usepackage{array}
\usepackage{tikz,pgfplots} 
\usepackage{tabularx}
\usepackage[linewidth=1pt, framemethod=tikz]{mdframed}
\usepackage{float}
\usepackage{hyperref} 
\usetikzlibrary{bayesnet} 
\newcolumntype{?}[1]{!{\vrule width #1}}

\usepackage{amssymb}

\journal{Mechanical Systems and Signal Processing}

\begin{document}

\begin{frontmatter}



\title{Inter-turbine Modelling of Wind-Farm Power using Multi-task Learning}



\author[inst1]{Simon M.\ Brealy\corref{cor1}}
\ead{sbrealy1@gmail.com}

\affiliation[inst1]{organization={Dynamics Research Group}, 
            addressline={Department of Mechanical Engineering, The University of Sheffield, Mappin Street}, 
            city={Sheffield},
            postcode={S1 3JD}, 
            country={UK}}

\author[inst2]{Lawrence A.\ Bull}

\affiliation[inst2]{organization={School of Mathematics and Statistics}, 
            addressline={University of Glasgow}, 
            city={Glasgow},
            postcode={G122 STA}, 
            country={Scotland}}

\author[inst3]{Pauline Beltrando}
\author[inst3]{Anders Sommer}

\affiliation[inst3]{organization={Vattenfall R\&D}, 
            addressline={Vattenfall AB}, 
            city={Älvkarleby},
            country={Sweden}}

\author[inst1]{Nikolaos Dervilis}
\author[inst1]{Keith Worden}

\cortext[cor1]{Corresponding author.}

\begin{abstract}
\noindent Because of the global need to increase power production from renewable energy resources, developments in the online monitoring of the associated infrastructure is of interest to reduce operation and maintenance costs. However, challenges exist for data-driven approaches to this problem, such as incomplete or limited histories of labelled damage-state data, operational and environmental variability, or the desire for the quantification of uncertainty to support risk management.

This work first introduces a probabilistic regression model for predicting wind-turbine power, which adjusts for wake-effects learned from data. Spatial correlations in the learned model parameters for different tasks (turbines) are then leveraged in a hierarchical Bayesian model (an approach to multi-task learning) to develop a ``metamodel'', which can be used to make power-predictions which adjust for turbine location---including on previously \textit{unobserved} turbines not included in the training data. The results show that the metamodel is able to outperform a series of benchmark models, while demonstrating a novel strategy for making efficient use of data for inference in populations of structures, in particular where correlations exist in the variable(s) of interest (such as those from wind-turbine wake-effects).
\end{abstract}


\end{frontmatter}


\section{Introduction}\label{sec:Introduction}


\subsection{Motivation}

Commitments to the expansion of the renewable energy sector equate to tripling the installed capacity globally by 2030 \cite{cop28_2023}. This growth presents some significant challenges, including how to efficiently manage the increasing cost of operations and maintenance (O\&M)---which represent approximately 40\% of the lifecycle cost of an offshore wind-farm \cite{noauthor_guide_2019}. Currently, O\&M is supported by online monitoring systems \cite{tautz-weinert_using_2017}, utilising sensors from across the turbines for decision-making; gaining the maximum possible insight from these data is therefore of interest to help maximise the economic viability of offshore wind-farms.

\subsection{Data-driven Modelling}

A common application of data for the online monitoring of wind-farms is in modelling the relationship between wind-speed and power (known as power-curve modelling). These models may be used as a damage-sensitive feature during operation, to inform the need for any remedial repairs \cite{mclean_physically_2023, rogers_2020}. As an additional benefit, these models can also be repurposed to make predictions of future wind-power generation (given forecasted weather inputs), which can help with risk mitigation in both operational and electricity market settings; this includes planning unit dispatch, maintenance scheduling and maximising profit in power trading \cite{Gilbert_2021,hanifi_2020}. A relatively recent review of the modelling approaches developed for power-curve monitoring is provided by Wang \textit{et al.}\ \cite{wang_approaches_2019}, which is a heavily researched area. However, in many cases the approaches used provide point, deterministic, predictions to predict output power. Inspection of typical power-curve data reveals that it is both heteroscedastic (the variance of the output power changes with wind-speed), and asymmetrically distributed. Deterministic approaches fail to capture these properties; as such, probabilistic approaches have gained greater attention more recently from researchers \cite{mclean_physically_2023, rogers_2020, Bazionis_2021}, which provide information on both the mean and the expected variance of the prediction. This additional information gives greater insight into model uncertainty, which could be helpful for managing risk in both engineering and commercial settings.

Further data-driven applications for the online monitoring of offshore wind-farms exist, including focussing on component health such as generator and gearbox-temperature monitoring, blade and bearing fault detection and, pitch and yaw misalignment \cite{stetco_machine_2019, maldonado-correa_using_2020, pandit_scada_2023}, or more system-level health e.g. detection of scour around the base of turbine monopiles \cite{jawalageri_data-driven_2024}. However, in these applications (and also more generally in the monitoring of engineering systems), significant challenges remain, including (1) a lack of labelled damage-state instances \cite{rytter_1993}, (2) short or incomplete histories of failure data \cite{yuan_machine_2020}, and (3) operational and environmental variability. These challenges can impede the level of sophistication, or even practical implementation of these systems \cite{farrar_structural_2012}.

\subsection{Multi-task Learning}

As a solution to the problem of limited data, multi-task learning (MTL)---a form of transfer learning (TL)---is a machine-learning approach where the goal is to learn multiple related tasks simultaneously, so that similarities (and differences) can be harnessed between tasks; in this way, model predictive accuracy is improved in domains where data are limited \cite{zhang_survey_2022,zhuang_comprehensive_2021,sun_2023}. In practice, a common approach to MTL is by the use of hierarchical Bayesian models, which are able to pool information across tasks via population-level parameters, while accounting for task-specific nuances via task-level parameters. By pooling information in this way, these models can better handle tasks with limited data, and produce robust probabilistic predictions. This model architecture can intuitively be applied to wind-farms, where individual turbines (tasks) may have individual differences (e.g.\ because of maintenance, damage or physical location), but are ultimately similar to each other as part of the wind-farm population. Hierarchical Bayesian models have recently been applied to other engineering infrastructure; \citet{bull_hierarchical_2022} uses a hierarchical Bayesian modelling approach to learn population-level and task-specific parameters in a combined inference, in the setting of a truck-fleet survival analysis. Similarly, \citet{dardeno_hierarchical_2024} and \citet{brealy_multitask_2024} use hierarchical Bayesian models, in order to transfer Frequency Response Function information between helicopter blades and wind-turbine foundation stiffness parameters respectively. In both cases, the uncertainty of parameters in data-sparse domains was reduced as a result of knowledge transfer between structures, as compared to modelling the data-sparse domains independently.

\subsection{Research Contribution}

In this work, a probabilistic power-curve model is first developed, that considers the heteroscedastic and asymmetric nature of power-curve data, using data from an in-service wind-farm. This model utilises wind-farm (population-level) wind-speed and direction features, with predictions indicating that a level of wind-directionality (from wind-turbine wake-effects), has been captured by the model. Spatial correlations found in the underlying model parameters inspired the incorporation of this model into a hierarchical Bayesian model architecture, where the population-level parameters were designed to capture spatial, inter-task correlations between turbines, which is described here as the ``metamodel''.

\begin{mdframed}
This model design has a tangible real-world benefit---if one were able to make accurate probabilistic predictions for a variable of interest, at a location without data (and perhaps even without associated telemetry), this may reduce the overall telemetry requirements among a population.
\end{mdframed}

The results show that for predicting wind-turbine output power, the metamodel outperforms a series of benchmark models, both in terms of mean prediction and predictive uncertainty, by making efficient use of data and accounting for environmental variability. The metamodel could also be described as a purely data-based approach to power modelling with adjustment for wake-effects, without the need for physics-based simulation (thus making it more efficient).

\subsection{Paper Layout}

The layout of the paper is as follows. Section \ref{section:dataset} describes the data used and the preprocessing steps. Sections \ref{section:bsbr} and \ref{section:meta_model} describe the models. Section \ref{section:model_training} describes the model training and scoring process. Results are analysed in Section \ref{section:results}. Conclusions and further work are described in Section \ref{section:conclusions}. Finally, links to the model code used in this study can be found in Section \ref{section:data_availability}.


\section{Dataset Overview and Preprocessing}\label{section:dataset}
\subsection{Description}
A Supervisory Control and Data Acquisition (SCADA) dataset was available to the authors of this work, which contains 224 wind-turbines, spanning three separate wind-farms between the years 2020-2023, at a ten-minute resolution. The measurements within come from a broad range of sensors, but generally fall into electrical, control, component/system temperature measurements, and ambient environmental parameters. This study focuses on one of these wind-farms (under the pseudonym ``\textit{Ciabatta}''), for the full year of 2021; this was chosen to maintain a full year of cyclical environmental variation, and to reduce computational overhead with a focus on demonstrating the methods herein.

The raw SCADA data, showing the power-curve relationship for the wind-turbines in wind-farm \textit{Ciabatta} are shown in blue in Figure \ref{fig:data_filtering}. In these data, a number of patterns typical to wind-farm operation are identifiable, including \textit{curtailments} (whereby turbine power is deliberately limited), \textit{shutdowns} (indicated by a power output of zero, despite significant wind-speeds), and \textit{power-boosting} (where power is raised above nominal turbine capacity). These three (controlled or otherwise) deviations of power away from \textit{normal} operation are typically related to technical and/or commercial constraints, dependent on factors which were considered outside the scope of this work. 

\subsection{Data Filtering}\label{section:filtering}
For the purposes of predicting power output during \textit{normal} operation, data most likely related to \textit{shutdowns} and \textit{curtailments} were filtered out. For the \textit{power-boosted} data, rather than filtering it, it was instead artificially reduced to the rated-power; this was done for two main reasons: (1) For some turbines, it accounted for a significant portion of the data at high wind-speeds, such that filtering it would create data-sparse regions for those turbines, making robust model development and comparison between different models more difficult. (2) From a practical perspective, if a model predicts that rated-power is achievable, and the specific conditions required for \textit{power-boosting} are known, then domain-experts could apply a manual correction if necessary. Additionally, since wake-effects are more prevalent at lower wind-speeds \cite{howland_wind_2019}, this modification was not expected to significantly interfere with wake-related behaviour in the data. A summary of the main criteria chosen to filter the data are shown in Table \ref{tab:filter_criteria}.

\newcolumntype{L}{>{\raggedright\arraybackslash}X}

\begin{table}[h]
  \centering
  \resizebox{\columnwidth}{!}{%
  \begin{tabularx}{\textwidth}{|L|L|L|}
  \hline
  \textbf{Filtering step} & \textbf{Description} & \textbf{Filtering target} \\ \hline
  1: Thresholding & Filter data based on blade pitch angle and rotor speed (RPM) & Curtailments and shutdowns \\ \hline
  2: Rate of change & Filter data where the RPM changes rapidly between time steps & Transient behaviour during startup and shutdown procedures \\ \hline
  3: Stationary data & Filter data where the power or wind-speed measurements are stationary & Curtailments and sensor faults \\ \hline
  4: Mahalanobis squared-distance outlier detection \cite{worden_damage_2000} & Filter data based on the distance in the blade pitch-angle vs.\ power relationship & Remaining outliers \\ \hline
  5: Standard-deviation outlier detection & Filter that retains the middle two standard deviations in power, for the binned \textit{freestream} wind-speed & Remaining outliers \\ \hline
  \end{tabularx}
  }
  \caption{Summary of the steps used to filter the raw SCADA data}
  \label{tab:filter_criteria}
  \end{table}

Between low and rated-power, in \textit{normal} operation the blade pitch angle is approximately zero, while during curtailment it increases to control the output power to a set level. When a turbine is \textit{offline} or during a \textit{shutdown}, the pitch angle is also higher than in \textit{normal} operation. Additionally, as the wind-speed increases beyond the cut-out wind-speed, the shaft speed (RPM) remains high, while the generated power drops considerably; these three sets of behaviours were targeted for filtering in step one, by defining set-thresholds on the blade-pitch-angle and rotor-speed, for differing levels of power. The second filtering step was introduced as it was found that in some cases, the RPM changed quickly; this was considered indicative of \textit{startup} and \textit{shutdown} procedures of the turbines, and not representative of \textit{normal} operation. In this step, data where the RPM increased or decreased by more than a set threshold, compared to the previous value were filtered out. In the data, instances of stationary wind-speed measurements were found, which were deemed likely to be due to sensor or SCADA system error. To combat this, step three was used to remove data where the wind-speed measurements were stationary in time. The same technique was also applied to remove any stationary power data that was not filtered by step one. Step four was used to help remove any remaining outliers in the pitch-angle vs.\ power relationship, by using a Mahalanobis Squared Distance (MSD)-based approach \cite{worden_damage_2000}, which takes account of both correlation and scale in the data. Finally, step five removed remaining outliers that were observed in the output power, by retaining only the middle two standard-deviations for a binned \textit{freestream} wind-speed---defined in Section \ref{section:features}. 

\begin{figure}[h]
  \centering
  \includegraphics[width=0.9\textwidth]{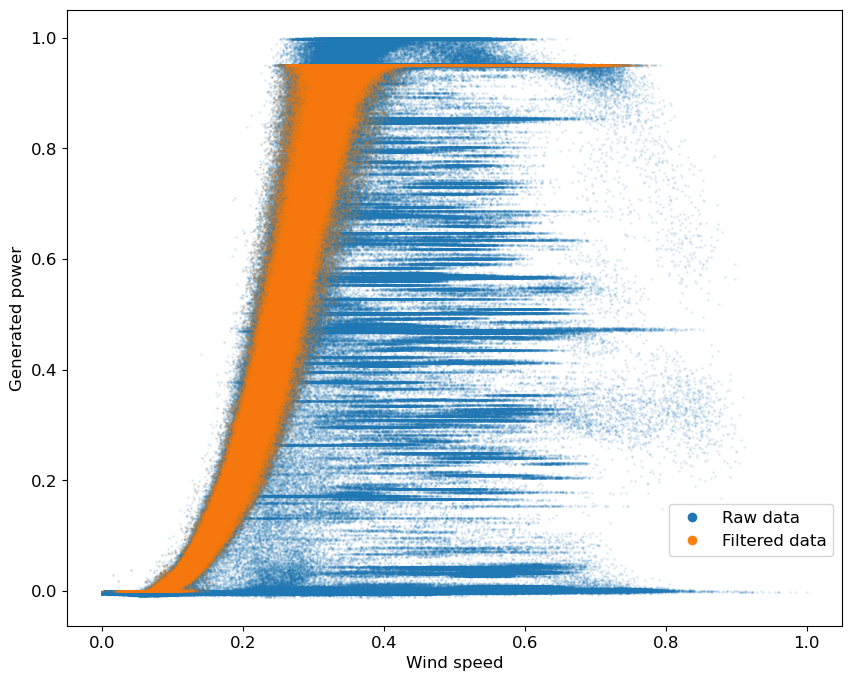}
  \caption{Data from all turbines in wind-farm \textit{Ciabatta} showing the power-curve relationship. Blue markers show the raw data prior to filtering, and orange markers show the data after filtering.}
  \label{fig:data_filtering}
\end{figure}

The filtered data are plotted in orange in Figure \ref{fig:data_filtering}, compared to the raw unfiltered data in blue. It can be seen that the filtered data appear much more representative of \textit{normal} operation only. 

\begin{figure}[h]
\centering
\begin{subfigure}{.5\textwidth}
  \centering
  \includegraphics[width=\textwidth]{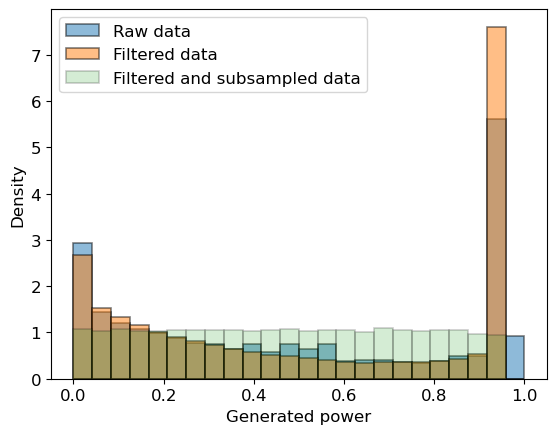}
  \caption{Generated Power.}
  \label{fig:power_hist}
\end{subfigure}%
\begin{subfigure}{.5\textwidth}
  \centering
  \includegraphics[width=\textwidth]{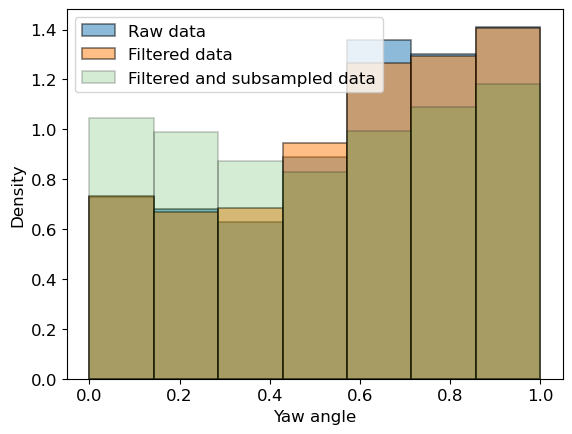}
  \caption{Yaw angle.}
  \label{fig:yaw_hist}
\end{subfigure}
\bigskip
\begin{subfigure}{.5\textwidth}
  \centering
  \includegraphics[width=\textwidth]{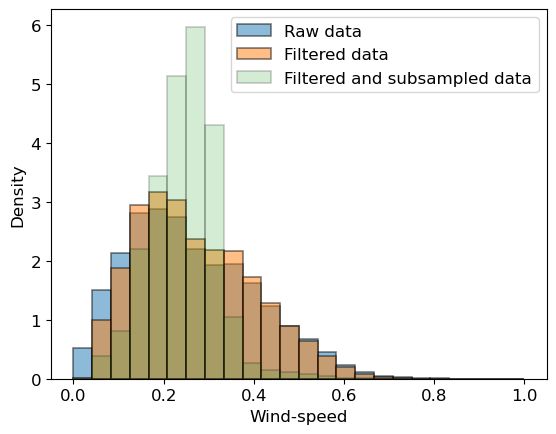}
  \caption{Wind-speed.}
  \label{fig:ws_hist}
\end{subfigure}
\caption{Density plots for the raw data (blue), filtered data (orange) and the subsampled data (green) for the generated power, yaw angle and wind-speed variables.}
\label{fig:filtering_histograms}
\end{figure}

To demonstrate how the filtering process influences the underlying distributions in the data, Figure \ref{fig:filtering_histograms} shows density plots for the generated power (\ref{fig:power_hist}), yaw angle (\ref{fig:yaw_hist}), and wind-speed (\ref{fig:ws_hist}) variables. The raw (unfiltered) data are shown in blue and the filtered data are shown in orange. Here it can be seen that in most cases the general shapes of the distributions after filtering are consistent. One clear difference can be seen at low (or zero) generated power; this is expected, since a large amount of data was present in this region and targeted for filtering. Some reductions in data density can be seen between 0.4--0.6 in generated power; this is because of the removal of the \textit{curtailment} data. Finally, there is an increase in the quantity of data in the final bin of the filtered generated power; this is because of the capping of the \textit{power-boosted} data. For completeness, the filtered data that had then undergone stratified (sub)sampling on both yaw angle and power features (discussed in Section \ref{section:meta_training}) are also shown here in green; in this case it can be seen that the desired effect of flattening the density plots for the yaw angle and power features had broadly been achieved.

Considering these filtered data (and power-curve data generally), there are two distinct features that present challenges for modelling: (1) the data are heteroscedastic---as the wind-speed changes, the variance in generated power changes significantly and (2), the skewness changes from being positive at low wind-speeds before cut-in, to negative above the rated wind-speed. These properties should be considered in the model design phase to ensure that the data are modelled appropriately.

\subsection{Model Training Features}\label{section:features}

For all models developed in this study, three features were used for predicting turbine-specific output power. First, came the freestream wind-speed (later referred to as Feature One); since data from a meteorological mast was not available, this was instead approximated as being equivalent to the maximum measured wind-speed of the turbines used in training, for a given timestamp. Since there are turbines on all edges of the wind-farm in the training data sets, this was deemed a reasonable approximation as measurements from ``unwaked'' turbines were always available. The authors are aware that anemometer sensor error will likely add noise to the data and could reduce model predictive performance as a result, however it is believed it does not detract from the novelty of the method developed in this work.

The second and third features were derived from what was considered a \textit{proxy} for wind-direction---the measured turbine yaw angle; these were assumed to be approximately equal to each other, since for operation in the \textit{normal} regime the control system should orientate the turbines in the direction of the incoming wind. Specifically, Features Two and Three took the sine and the cosine of the yaw angle; this was done to create a smooth transition in the inputs between the equivalent angles of zero and 360 degrees. The median values were used for these features which were taken across the turbines for a given timestamp, to obtain an \textit{averaged} wind-direction, less subject to individual turbine noise. Finally, all features were scaled between zero and one for both numerical efficiency, and data anonymity.

The motivations for choosing these features were as follows: (1) using the freestream wind-speed in combination with wind-directional features, enables the possibility to capture some of the wake-effects that may be present in the data. (2) While not the main focus of this study, population-level features are more representative of forecast data by data providers such as ECMWF \cite{ecmwf_2024}, since higher resolution turbine-specific forecasts that also account for turbine wake-effects are not readily available.


\section{B-spline Beta Regression Model}\label{section:bsbr}

The heteroscedacity, skewness and bounded nature of power-curve data share similarities with the shape of the beta distribution, which is bounded between zero and one, and whose variance changes with the mean of the distribution. These two factors motivated the development of the B-Spline Beta Regression Model (BSBR), which is a probabilistic regression-based model utilising a beta-likelihood. \citet{mclean_physically_2023} similarly used a beta-likelihood, and the description of the BSBR model that follows is very similar to that of \citet{capelletti_wind_2024} whom also used a spline-based approach.

\subsection{Generalised Linear Models}
Suppose one wishes to model the distribution of a real target variable $\mathbf{y} = [y_1, y_2 ... y_N]^T$, given an input design matrix $\mathbf{X}$. Generalised Linear Models (GLMs) are a generalisation of linear regression, which are characterised by three components: 

\begin{enumerate}
  \item a distribution for modelling $\mathbf{y}$ (which must be from the exponential family of distributions);
  \item a linear predictor $f = \mathbf{X}\boldsymbol{\beta}$, where $\boldsymbol{\beta}$ is a vector of regression parameters;
  \item a link function $g(\cdot)$ such that $E(\mathbf{y}|\mathbf{X}) = \boldsymbol{\mu} = g^{-1} (f)$ which allows linear models to be related to a real target variable, $\mathbf{y}$, via the link function.
\end{enumerate}

\subsection{Beta Regression Model}\label{sec:beta_regression_model}

Beta regression is one such type of GLM, where the beta distribution is chosen for modelling $\mathbf{y}$. Beta distributions are bounded in the interval between zero and one, while their variance can change with the mean of the distribution; this has clear similarities with power-curve data, which can be sensibly bounded between zero and rated power, with trivial rescaling. 
The beta distribution for a scalar $y$ is defined as,

\begin{equation}\label{eq:beta_dist}
  \begin{gathered}
  f(y|\mu,\phi) =\frac{\Gamma(a+b)}{\Gamma(a) \Gamma(b)} y^{(a-1)}(1-y)^{(b-1)} \\
      a = \mu \phi, \quad b = (1 - \mu)\phi
  \end{gathered}
\end{equation}
\noindent where $\Gamma(\cdot)$ denotes the gamma function, and where $0<\mu<1$ and $\phi>0$. The mean and variance of $y$ are,

\begin{equation}
\mathbb{E}(y) =\mu
\end{equation}
  
\begin{equation}
\mathbb{V}(y) =\frac{\mu(1-\mu)}{(1 + \phi)}
\end{equation}
\noindent whereby $\mu$ can be described as the mean parameter, and $\phi$ the precision parameter since it is inversely proportional to the variance. 

In this case for modelling power-curve data, consider a linear predictor model, $f_1 = \mathbf{X}\boldsymbol{\eta}$ for a GLM, where $\boldsymbol{\eta}$ is a vector of regression parameters. 
The link function is chosen to be the \textit{logit} function (equivalent to $g^{-1}$ equalling the \textit{expit} function \cite{verhulst1845research}), such that,

\begin{equation}\label{eq:mu}
  \boldsymbol{\mu} = expit(\mathbf{X}\boldsymbol{\eta})
\end{equation}

\noindent where $\boldsymbol{\mu}$ is the mean parameter vector for the beta distribution. The choice of the \textit{logit} link function ensures that $0<\mathbb{E}(\mathbf{y})<1$, which in words, means the predictive mean of the model must lie between zero and one, intuitively matching the behaviour of \textit{normalised} power-curve data. For additional flexibility in the model, the precision parameter vector $\boldsymbol{\phi}$ (in the beta distribution) was also allowed to vary with $\mathbf{X}$ according to an additional linear model $f_2 = \mathbf{X}\boldsymbol{\zeta}$, where $\boldsymbol{\zeta}$ is an additional vector of regression parameters. Since $\boldsymbol{\zeta}$ must be non-negative (to ensure that $\mathbb{V}(\mathbf{y}$) is positive), a natural \textit{logarithm} link function is chosen (equivalent to $g^{-1}$ equalling the \textit{exponential} function), such that, 

\begin{equation}\label{eq:phi}
  \boldsymbol{\phi} = exp(\mathbf{X}\boldsymbol{\zeta})
\end{equation}

\noindent This model definition allows both the mean, $\boldsymbol{\mu}$, and precision, $\boldsymbol{\phi}$, parameter vector entries to vary as a function of $\mathbf{X}$, resulting in beta distributions whose shapes vary with the input feature(s); this is similar to the ``Variable dispersion beta regression model'' described in \cite{capelletti_wind_2024}.  

\subsection{B-spline Linear Models}
\begin{figure}[h]
  \centering
  \includegraphics[width=0.8\textwidth]{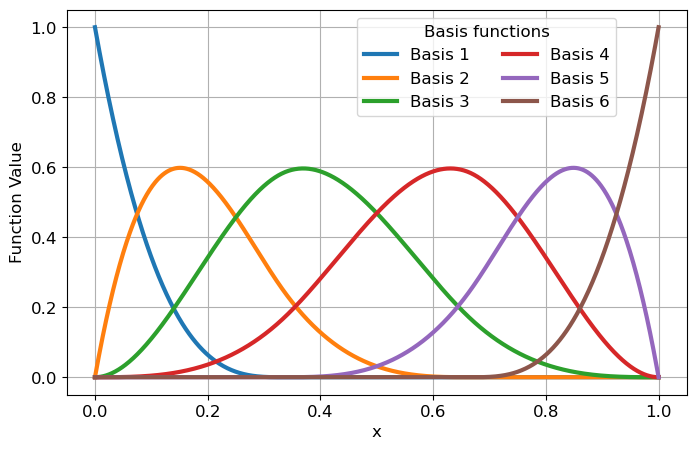}
  \caption{B-spline basis-functions over the interval of zero and one, with four interior knots.}
  \label{fig:bsplines}
\end{figure}

Regression spline models are chosen for functions $f_1$ and $f_2$; these provide greater flexibility in modelling nonlinear outputs compared to polynomial regression \cite{James2013}, while still being linear in parameters. Regression splines are piece-wise continuous polynomial functions, joined at \textit{knots}, where the first and second derivatives of adjacent functions are equal; this ensures a smooth transition between polynomials. In practice, polynomials of degree three (cubic splines), balancing smoothness and flexibility, are most commonly used \cite{deboor_1978} and are chosen here. Specifically, the B-spline basis-function approach is used for its numerical stability and computational efficiency \cite{piegl_nurbs_1996}. Figure \ref{fig:bsplines} shows the B-spline basis-functions for polynomials of degree three, using four interior knots between zero and one. Parameters (weights) are learned for each basis-function during model fitting, with model predictions calculated as a weighted sum of parameters and basis-functions.

\begin{figure}[h]
  \centering
  \includegraphics[width=0.8\textwidth]{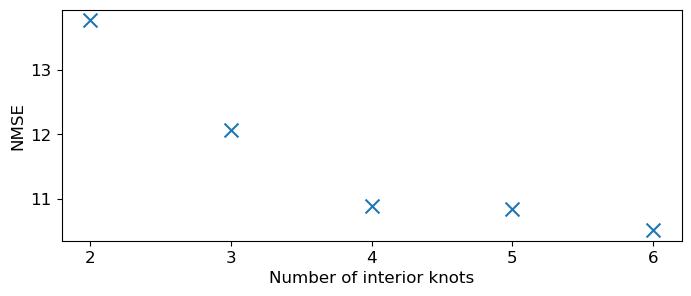}
  \caption{Comparison of NMSE score vs. the number of (evenly-spaced) interior knots per feature, for the no-pooling BSBR model described in Section \ref{section:bsbr_demo}.}
  \label{fig:knot_comparison}
\end{figure}

Using the B-spline basis-function approach, the design matrix $\mathbf{X}$ was constructed and used alongside the linear model parameters $\boldsymbol{\eta}$ and $\boldsymbol{\zeta}$ for the mean ($\boldsymbol{\mu}$) and precision ($\boldsymbol{\phi}$) parameters. B-splines of order four were chosen, with four, uniformly spaced, interior knots per feature. Uniform spacing was chosen for simplicity, while additional knots resulted in no significant improvement in the \textit{normalised} mean squared error (NMSE) (defined in Section \ref{section:scoring}); this is demonstrated in Figure \ref{fig:knot_comparison}. Further optimisation of the number of knots for individual features, and the location of knots was not considered, and is left as an area for further model refinement. 

\subsection{Graphical Model}

A graphical representation of the B-spline beta regression (BSBR) model is displayed in Figure \ref{fig:graphical_bsbr}. Latent variables (to be estimated) are depicted as unshaded circled nodes, while the shaded circled node represents the observed variable. The dotted lines indicate parameters linked by a transformation, while the arrows indicate that the target parameter is drawn from a distribution using the preceding parameters. The plate represents multiple instances of the contained node.

\begin{figure}[h]
  \centering
  \begin{tikzpicture}
  \node[obs]                                     (y)  {$y_{i,t}$};
  \node[latent, above=0.5cm of y, xshift=-2.5cm] (mu)  {$\boldsymbol{\mu}_{t}$};
  \node[latent, below=0.5cm of y, xshift=-2.5cm] (phi)  {$\boldsymbol{\phi}_{t}$};
  \node[latent, left=1cm of mu]                   (eta) {$\boldsymbol{\eta}_t$};
  \node[latent, left=1cm of phi]                   (zeta) {$\boldsymbol{\zeta}_t$};

  \edge {mu,phi} {y} ; %

  \draw[dotted] (eta) -- (mu) ; %
  \draw[dotted] (zeta) -- (mu) ; %
  \draw[dotted] (eta) -- (phi) ; %
  \draw[dotted] (zeta) -- (phi) ; %

  \plate [inner sep = .3cm]{N} {(y)} {$i\in1:N_t$} ;

  \end{tikzpicture}
  \caption{A graphical representation of the B-spline beta regression model.}
  \label{fig:graphical_bsbr}
\end{figure}

Here, $\textit{t}$ relates to the training turbine, while $\textit{N}_t$ relates to the number of training data points for that turbine. In this initial case, independent models are learned per turbine \textit{t}, used to model the observed target variable $y_{i,t}$. The column vector $\mathbf{y}_t$, with entries \( y_{i,t} \) for \( i = 1, 2, \ldots, N_t \), is distributed according to the reparameterised beta distribution,

\begin{equation} \label{eq:y}
  \mathbf{y}_t \sim \mathrm{Beta}(\boldsymbol{\mu}_{t},\boldsymbol{\phi}_{t})
\end{equation}

\noindent Additionally, \textit{hyperpriors} were placed over the estimated parameters $\boldsymbol{\eta}_t$ and $\boldsymbol{\zeta}_t$, according to a Normal distribution,

\begin{equation} \label{eq:eta}
  \{\boldsymbol{\eta_t}, \boldsymbol{\zeta_t}\} \sim \mathcal{N}(0, 3) \quad \text{(element-wise)}
\end{equation}

\noindent where the means and standard deviations of 0 and 3 respectively were found to be weakly informative, while sufficiently aiding convergence.

\subsection{Inference}
The observed variables in graphical models may be referred to as evidence nodes \cite{gelman2013bayesian}. In the example in Figure \ref{fig:graphical_bsbr}, the regression would have the following evidence nodes:

\begin{equation}
\mathcal{E}=\left\{\mathbf{y}_t\right\}
\end{equation}

\noindent Latent variables on the other hand are \textit{hidden} nodes,

\begin{equation}
\mathcal{H}=\left\{\boldsymbol{\eta}_t, \boldsymbol{\zeta}_t\right\}
\end{equation}

\noindent Bayesian inference relies on finding the posterior distribution of $\mathcal{H}$ given $\mathcal{E}$, that is, the distribution of the unknown parameters given the data,

\begin{equation}
  \begin{aligned}
    p(\mathcal{H} \mid \mathcal{E})=\frac{p(\mathcal{E} \mid \mathcal{H})p(\mathcal{H})}{p(\mathcal{E})}\\
    & = \frac{p(\mathbf{y_t} \mid \boldsymbol{\eta}_t, \boldsymbol{\zeta}_t) p(\boldsymbol{\eta}_t) p(\boldsymbol{\zeta}_t)}{p(\mathbf{y_t})}\\
    & = \frac{p(\mathbf{y_t} \mid \boldsymbol{\eta}_t, \boldsymbol{\zeta}_t) p(\boldsymbol{\eta}_t) p(\boldsymbol{\zeta}_t)}{\int\int p(\mathbf{y_t} \mid \boldsymbol{\eta}_t, \boldsymbol{\zeta}_t)p(\boldsymbol{\eta}_t) p(\boldsymbol{\zeta}_t) d\boldsymbol{\eta}_t d\boldsymbol{\zeta}_t}
  \end{aligned}
  \label{eq:bsbr_posterior}
\end{equation}

As is common in Bayesian inference problems, the integral in this case is intractable, and cannot be solved analytically. In this work, the posterior distribution is approximated using the Monte Carlo Markov Chain (MCMC) method, via the no U-turn (NUTS) implementation of Hamiltonian Monte Carlo (HMC) \cite{homan_no-u-turn_2014}.  

\section{A Metamodelling Approach}\label{section:meta_model}

\subsection{Hierarchical Bayesian Modelling for Multi-task Learning}

In Section \ref{section:bsbr}, the presented BSBR model can be described as being an example of single task learning (STL), since the task-level parameters $\boldsymbol{\eta}_t$ and $\boldsymbol{\zeta}_t$ are learned independently for each task; in this case, there is no-pooling (NP) of information between tasks. By contrast, a model containing parameters solely at the population-level which are shared across all tasks, can be said to have a complete-pooling (CP) modelling structure. In this case, all tasks are treated as identical, which can lead to poor model generalisation to new tasks. Alternatively, hierarchical Bayesian models (also known as multi-level models) provide a middle-ground, containing both task-specific parameters (allowing for differences between population members), and population-level parameters, which encourage task parameters to be similar; this modelling structure can be described as having partial-pooling (PP) of information. These models are often used as an approach to multi-task learning (MTL), where the goal is to improve the performance and generalisation of a model, by training it on multiple related tasks simultaneously \cite{murphy2012machine}.

In the context of power-curve modelling, allowing PP of information makes intuitive sense; wind-turbines within a wind-farm are nominally identical, so they are likely to have similar power-curves (at a population-level), but differences may also arise from varying wake patterns at individual turbine locations (the task-level). An additional benefit of PP of information, is that data-poor tasks in particular are able to borrow statistical strength from data-rich tasks. While in this context data availability is not a significant problem, it is common in the field of data-driven SHM \cite{farrar_structural_2012}, which has led to these modelling approaches being used as a solution in the domain of population-based SHM (PBSHM) research \cite{bull_hierarchical_2022,dardeno_hierarchical_2024}.

\subsection{Metamodel Design}
In the results in Section \ref{section:bsbr_results}, correlations are observed among the learned $\boldsymbol{\eta}_t$ and $\boldsymbol{\zeta}_t$ parameters of the NP BSBR models, in relation to the turbine $x$ and $y$ spatial coordinates. This fact motivated the integration of a metamodel, which can be described as a model over models. The metamodel was designed to capture the inter-turbine relationships in power-curve parameters, which could be used to predict power for previously unobserved turbines, not necessarily in the training data. This model design has a tangible real-world benefit---if one were able to make accurate probabilistic predictions for a variable of interest, at a location without data (and perhaps even without associated telemetry), this may reduce the overall telemetry requirements among a population. While in the use case here data are readily available at each turbine location, it serves as a useful testbed to experiment with the modelling approach.

First-order linear metamodels were chosen here for both simplicity and to reduce the risk of overfitting to the (potentially noisy) training data. In matrix form, the parameter vectors $\boldsymbol{\eta}$ and $\boldsymbol{\zeta}$ are described via the linear metamodels,

\begin{equation}\label{eq:eta_meta}
  \boldsymbol{\eta} = \mathbf{C_x}\mathbf{M}_{\boldsymbol{\eta}_\mathbf{x}} + \mathbf{C_y}\mathbf{M}_{\boldsymbol{\eta}_\mathbf{y}}
\end{equation}

\begin{equation}\label{eq:zeta_meta}
  \boldsymbol{\zeta} = \mathbf{C_x}\mathbf{M}_{\boldsymbol{\zeta}_\mathbf{x}} + \mathbf{C_y}\mathbf{M}_{\boldsymbol{\zeta}_\mathbf{y}}
\end{equation}

\noindent where $\mathbf{C_x}$ and $\mathbf{C_y}$ are the linear regression design matrices for the normalised $x$ and $y$ coordinates for each turbine, and $\mathbf{M}_{\boldsymbol{\eta}_\mathbf{x}}$ and $\mathbf{M}_{\boldsymbol{\eta}_\mathbf{y}}$ are the metamodel parameter matrices for parameter vector $\boldsymbol{\eta}$. Similarly, $\mathbf{M}_{\boldsymbol{\zeta}_\mathbf{x}}$ and $\mathbf{M}_{\boldsymbol{\zeta}_\mathbf{y}}$ are the metamodel parameter matrices for parameter vector $\boldsymbol{\zeta}$.

As in the NP BSBR model, \textit{hyperpriors} were used to aid model convergence, and were placed over the metamodel parameter matrices assuming normal distributions, 

\begin{equation}\label{eq:meta_priors}
  \{\mathbf{M}_{\boldsymbol{\eta}}, \mathbf{M}_{\boldsymbol{\zeta}}\} \sim \mathcal{N}(0, 3) \quad \text{(element-wise)}
\end{equation}

\noindent where $\mathbf{M}_{\boldsymbol{\eta}} = \left\{\mathbf{M}_{\boldsymbol{\eta}_\mathbf{x}},\mathbf{M}_{\boldsymbol{\eta}_\mathbf{y}}\right\}$ and $\mathbf{M}_{\boldsymbol{\zeta}} = \left\{\mathbf{M}_{\boldsymbol{\zeta}_\mathbf{x}},\mathbf{M}_{\boldsymbol{\zeta}_\mathbf{y}}\right\}$ and the \textit{prior} means and standard deviations were set equal to 0 and 3 respectively, given the distribution of learned parameters seen for the individual NP BSBR models; these were again found to be weakly informative, while sufficiently aiding convergence. Figure \ref{fig:graphical_meta} shows a graphical representation of the model. As with Figure \ref{fig:graphical_bsbr}, arrows represent distribution inputs, while dotted lines show calculated inputs, with plates indicating multiple instances of their contained nodes.

\begin{figure}[H]
  \centering
  \begin{tikzpicture}
  \node[obs]                                     (y)  {$y_{i,t}$};
  \node[latent, above=0.6cm of y, xshift=-2.5cm] (mu)  {$\boldsymbol{\mu}_{t}$};
  \node[latent, below=0.6cm of y, xshift=-2.5cm] (phi)  {$\boldsymbol{\phi}_{t}$};
  \node[latent, left=1cm of mu]                   (eta) {$\boldsymbol{\eta}_t$};
  \node[latent, left=1cm of phi]                   (zeta) {$\boldsymbol{\zeta}_t$};
  \node[latent, left=1cm of eta] (meta_eta) {$\mathbf{M}_{\boldsymbol{\eta}}$};
  \node[latent, left=1cm of zeta] (meta_zeta) {$\mathbf{M}_{\boldsymbol{\zeta}}$};

  \draw[dotted] (meta_eta) -- (eta) ; %
  \draw[dotted] (meta_zeta) -- (zeta) ; %
  \draw[dotted] (eta) -- (mu) ; %
  \draw[dotted] (zeta) -- (mu) ; %
  \draw[dotted] (eta) -- (phi) ; %
  \draw[dotted] (zeta) -- (phi) ; %

  \edge {mu} {y} ;
  \edge {phi} {y} ;
  
  \plate [inner sep = .3cm]{N} {(y)} {$i\in1:N_t$} ;
  \plate [inner sep = .3cm]{T} {(y)(mu)(phi)(eta)(zeta)(N.north west)(N.south east)} {$t\in1:T$} ;

  \end{tikzpicture}
  \caption{A graphical representation of the metamodel. Arrows leading from parameters indicate they are distribution parameters, whereas dotted lines indicate parameters are associated via a transformation.}
  \label{fig:graphical_meta}
\end{figure}

\noindent Note that the metamodel parameters are located outside the plates, and are therefore learned at the population-level, while the parameters $\boldsymbol{\eta}_t$ and $\boldsymbol{\zeta}_t$ are determined specifically for each turbine. 

\subsection{Metamodel Predictions}


\begin{mdframed}
  Following posterior estimation of the population-level metamodel parameters, one can make probabilistic predictions at locations of interest, based on spatial coordinates alone, 

  \begin{equation}\label{eq:meta_eta_pred}
    \boldsymbol{\eta^*} = \mathbf{C_x^*}\mathbf{\hat{M}}_{\boldsymbol{\eta}_\mathbf{x}} + \mathbf{C_y^*}\mathbf{\hat{M}}_{\boldsymbol{\eta}_\mathbf{y}}
  \end{equation}
  
  \begin{equation}\label{eq:meta_zeta_pred}
    \boldsymbol{\zeta^*} = \mathbf{C_x^*}\mathbf{\hat{M}}_{\boldsymbol{\zeta}_\mathbf{x}} + \mathbf{C_y^*}\mathbf{\hat{M}}_{\boldsymbol{\zeta}_\mathbf{y}}
  \end{equation}
\end{mdframed}

\noindent where $\boldsymbol{\eta^*}$ and $\boldsymbol{\zeta^*}$ are the predicted metamodel parameters for all turbine locations of interest, $\mathbf{{C_x}^*}$ and $\mathbf{{C_y}^*}$ are their associated first-order linear regression design matrices for the normalised spatial coordinates, and $\mathbf{\hat{M}}_{\boldsymbol{\eta}_\mathbf{x}}$, $\mathbf{\hat{M}}_{\boldsymbol{\eta}_\mathbf{y}}$, $\mathbf{\hat{M}}_{\boldsymbol{\zeta}_\mathbf{x}}$ and $\mathbf{\hat{M}}_{\boldsymbol{\zeta}_\mathbf{y}}$ are the estimated metamodel parameters. The predicted parameters can then be used in line with the population-level input features, $\mathbf{X}$, to make probabilistic power-predictions.

\subsection{Metamodel Benchmarking}
The model presented in Section \ref{section:meta_model} is a hierarchical Bayesian model, incorporating a metamodel---this model is simply referred to as the metamodel. To help assess the performance of the metamodel, it is benchmarked against three models which have no metamodel structure or spatial component. The first of these, is the NP BSBR model that was first introduced, which treats each turbine independently---i.e.\ there is no-pooling of information between turbines, with separate sets of parameters for each one. Secondly the metamodel is compared to a partially-pooled (PP) hierarchical model (shown in Figure \ref{fig:pp_model}), where the turbine-specific parameters are drawn from shared population-level distributions. For completeness, the metamodel is compared to a complete-pooling (CP) approach, where all turbines are considered identical and a single set of model parameters are learned for all turbines; this is shown in Figure {\ref{fig:cp_model}}. In the PP model, the $\eta$ and $\zeta$ parameter vectors were assumed to be normally distributed, controlled by the population-level distribution parameters $\left\{\boldsymbol{\mu}_{\eta}, \boldsymbol{\sigma}_{\eta}, \boldsymbol{\mu}_{\zeta}, \boldsymbol{\sigma}_{\zeta}\right\}$. For further details, including the details of the \textit{hyperpriors} placed over the parameters in the models, \texttt{stan} code containing the four models is provided in the links in Section \ref{section:data_availability}.

\begin{figure}[h]
  \centering
  \begin{tikzpicture}
  \node[obs]                                     (y)  {$y_{i,t}$};
  \node[latent, above=0.8cm of y, xshift=-2.5cm] (mu)  {$\boldsymbol{\mu}_{t}$};
  \node[latent, below=0.8cm of y, xshift=-2.5cm] (phi)  {$\boldsymbol{\phi}_{t}$};
  \node[latent, left=1cm of mu]                   (eta) {$\boldsymbol{\eta}_t$};
  \node[latent, left=1cm of phi]                   (zeta) {$\boldsymbol{\zeta}_t$};
  \node[latent, above=0.2cm of eta, xshift=-2cm] (mu_eta) {$\boldsymbol{\mu_{{\eta}}}$};
  \node[latent, below=0.2cm of eta, xshift=-2cm] (sig_eta) {$\boldsymbol{\sigma_{{\eta}}}$};
  \node[latent, above=0.2cm of zeta, xshift=-2cm] (mu_zeta) {$\boldsymbol{\mu_{{\zeta}}}$};
  \node[latent, below=0.2cm of zeta, xshift=-2cm] (sig_zeta) {$\boldsymbol{\sigma_{{\zeta}}}$};


  \draw[dotted] (eta) -- (mu) ; %
  \draw[dotted] (zeta) -- (mu) ; %
  \draw[dotted] (eta) -- (phi) ; %
  \draw[dotted] (zeta) -- (phi) ; %

  \edge {mu_eta} {eta} ; %
  \edge {sig_eta} {eta} ; %
  \edge {mu_zeta} {zeta} ; %
  \edge {sig_zeta} {zeta} ; %

  \edge {mu} {y} ;
  \edge {phi} {y} ;
  
  \plate [inner sep = .3cm]{N} {(y)} {$i\in1:N_t$} ;
  \plate [inner sep = .3cm]{T} {(y)(mu)(phi)(eta)(zeta)(N.north west)(N.south east)} {$t\in1:T$} ;

  \end{tikzpicture}
  \caption{Partially-pooled (PP) BSBR graphical model.}
  \label{fig:pp_model}
\end{figure}


\begin{figure}[h]
  \centering
  \begin{tikzpicture}
  \node[obs]                                     (y)  {$y_{i,t}$};
  \node[latent, above=0.5cm of y, xshift=-2.5cm] (mu)  {$\boldsymbol{\mu}$};
  \node[latent, below=0.5cm of y, xshift=-2.5cm] (phi)  {$\boldsymbol{\phi}$};
  \node[latent, left=1cm of mu]                   (eta) {$\boldsymbol{\eta}$};
  \node[latent, left=1cm of phi]                   (zeta) {$\boldsymbol{\zeta}$};

  \edge {mu,phi} {y} ; %

  \draw[dotted] (eta) -- (mu) ; %
  \draw[dotted] (zeta) -- (mu) ; %
  \draw[dotted] (eta) -- (phi) ; %
  \draw[dotted] (zeta) -- (phi) ; %

  \plate [inner sep = .3cm]{N} {(y)} {$i\in1:N$} ;

  \end{tikzpicture}
  \caption{Completely-pooled (CP) BSBR graphical model.}
  \label{fig:cp_model}
\end{figure}

\section{Model Training and Testing}\label{section:model_training}

\subsection{Training and Testing Datasets}\label{section:traintest}

\subsubsection{BSBR Model Demonstration}\label{section:bsbr_demo}
In the first instance, to demonstrate the suitability of the BSBR model for capturing the heteroscedastic and asymmetric nature of power-curves, the model is trained and tested on individual turbines (i.e.\ using the no-pooling model definition). A subset of the filtered data that had undergone stratified sampling by yaw angle is used, to provide the models with an approximately even distribution of data across yaw angles, to try to help make the models more generalisable to all wind-directions. The training and testing datasets consisted of 4000 and 1000 data points per turbine respectively.

\subsubsection{Metamodel}\label{section:meta_training}
In the SCADA data, it was noted that there were significant proportions at zero or maximum power; because of this, additional stratified sampling was applied based on measured turbine output power. The reasons for doing this were twofold: (1) to train and assess models equally for all regions of the power-curve (2) because of the heteroscedastic nature of the power-curve, harmonising the proportions of data in different regions of the power-curve between turbines was considered to make comparisons between predictive metrics more reasonable. Since the metamodel aims to capture spatial correlations induced by wake-effects, reason (1) is particularly important, since wake-effects are more prevalent in the transitional region of the power-curve \cite{howland_wind_2019}, which is comparatively data-sparse. The distributions of the resulting (sub)sampled data are shown in Figures \ref{fig:power_hist} and \ref{fig:yaw_hist}.

\begin{figure}[h]
  \centering
    \includegraphics[width=0.8\textwidth]{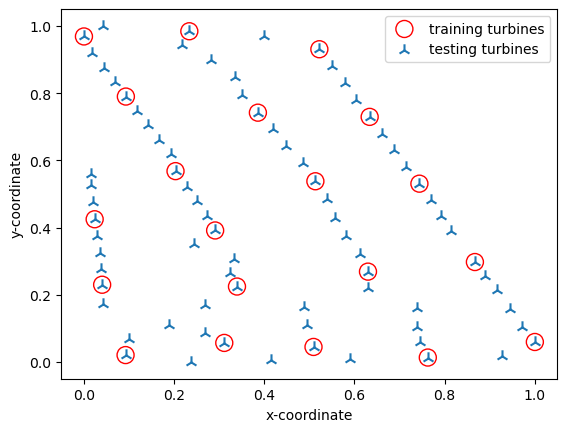}
    \caption{Map showing the training and testing turbines for the metamodel. Red circles show the locations of the turbines used in training, while all turbines (shown by the blue markers) were used in testing.}
    \label{fig:train_test_map}
  \end{figure}

Additionally, for the metamodel, since a single model is learned for all data simultaneously, computational expense is increased significantly, so a reduced number of training points were needed. The resulting training and testing sets, having undergone stratified sampling on both the yaw angle and output power, consisted of just 250 points per turbine, which were used for the metamodel and benchmark models. Furthermore, since the metamodel and benchmark models are to be trained on a subset of the turbines, as an example, one quarter of the turbines were used, which were approximately equally-spaced throughout the wind-farm. The chosen training and the testing turbines are highlighted in Figure \ref{fig:train_test_map}. The resulting combined training and test sets consisted of 5000 and 20000 data points respectively.

\subsection{Model Training}\label{sec:model_training}

As previously discussed, MCMC is used to estimate the posterior distributions of our parameters, given observed levels of power per turbine. For all the models in this study, MCMC was implemented using \texttt{stan} \cite{stan_development_team_stan_2024}, with 2000 burn-in and 2000 inference samples per chain, and four chains. In all cases, the \textit{maximum} \textit{$\hat{R}$} convergence statistics did not exceed 1.01, which suggests proper mixing of the chains \cite{vehtari_rank-normalization_2021}. Additionally, the \textit{minimum} effective sample sizes (ESS) were greater than 400, a benchmark considered suitable for stable parameter estimates and proper interpretation of the \textit{$\hat{R}$} statistics \cite{vehtari_rank-normalization_2021}.

\subsection{Model Scoring}\label{section:scoring}
In order to quantify the success of each of the models, two performance metrics were used. Firstly, the normalised mean squared error (NMSE) assesses the goodness-of-fit, or how close point predictions are to their test value. The NMSE functions similarly to percentage error, where a score of zero indicates a perfect model, while a score of 100 is equivalent to a model simply predicting the mean value of the data. NMSE is defined as,

\begin{equation}\label{eq:NMSE}
  \mathrm{NMSE}=\frac{100}{N \sigma_y^2} \sqrt{(\mathbf{y}-\widehat{\mathbf{y}}){^\intercal}(\mathbf{y}-\widehat{\mathbf{y}})}
\end{equation}
\noindent where $N$ is the number of data points, $\sigma_y^2$ is the variance of the measured data, $\mathbf{y}$ the measured data, and $\widehat{\mathbf{y}}$ the model predictions. 

Since the models are probabilistic and include predictive variance, the joint-log-likelihood (JLL) is also calculated as a measure of how well the uncertainty in the measured data was captured by the model. The JLL is calculated as,

\begin{equation}\label{eq:JLL}
  JLL=\frac{\sum_{s=1}^{S}\sum_{i=1}^{N_t}\log{f(y_{{i,t}_s}|{\mu}_{{i,t}_s}, {\phi}_{{i,t}_s})}}{S}
\end{equation}

\noindent where $f(y_{{i,t}_s}|{\mu}_{{i,t}_s}, {\phi}_{{i,t}_s})$ is the probability density function of the beta distribution as defined in equation (\ref{eq:beta_dist}), and \textit{S} is the number of posterior samples. The larger the JLL, the better the model has captured the true distribution of the test data. Given the prevalence of power-curve modelling in the literature, it is worth stating that since the model metrics are dependant upon both the training features and test data used, they are best used as a comparison of the models within this paper only.


\section{Results and Discussion}\label{section:results}

\subsection{BSBR Model Demonstration}\label{section:bsbr_results}

\begin{figure}[h!]
  \centering
  \begin{subfigure}{0.6\textwidth}
    \includegraphics[width=\textwidth]{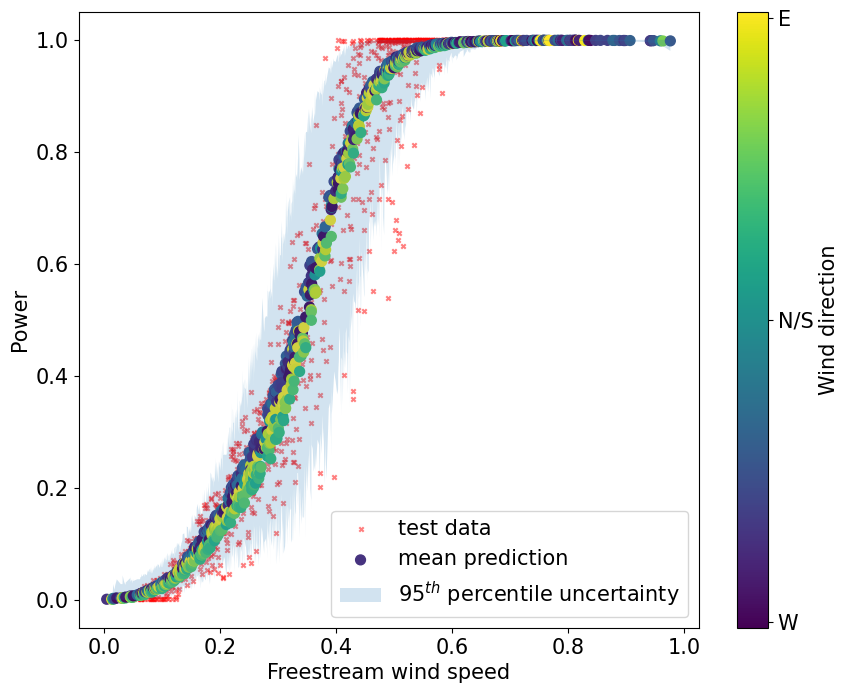}
    \caption{West-side turbine.}
    \label{fig:west_side}
  \end{subfigure}%

  \bigskip
  \begin{subfigure}{0.6\textwidth}
    \includegraphics[width=\textwidth]{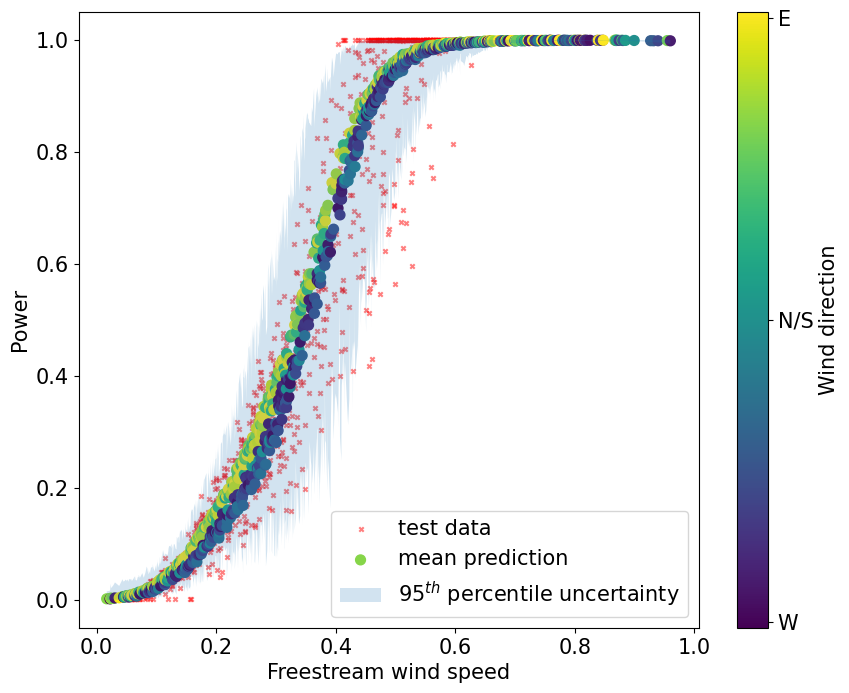}
    \caption{East-side turbine.}
    \label{fig:east_side}
  \end{subfigure}
  \caption{NP BSBR model predictions for two separate turbines. Test data (1000 points) are shown by red crosses, mean predictions are shown as circles, coloured by the wind-direction. The blue shaded areas show the 95$^{th}$-percentile of the predictive variance.}
  \label{fig:BSBR model predictions}
  \end{figure}

To first demonstrate probabilistic predictions that can be made with the BSBR models (with no-pooling), Figure \ref{fig:BSBR model predictions} shows predictions from models of turbines on both the west and east-side of the wind-farm in subplots (a) and (b) respectively, using the 1000 testing data points in each case. Test data are indicated by red crosses, model mean predictions are indicated by circles (which are coloured by the value of the wind-direction), and model uncertainty is represented by the blue shaded area--indicating the 95$^{th}$-percentile of predictions.

There are a number of interesting observations. Firstly, the predictive uncertainty appears to match the data, with reduced variance at the lower and upper ends of the curve, and increased variance in the transitional region between turbine minimum and maximum power. Approximately 5\% of the test points also lie outside the 95$^{th}$-percentile of the predictive variance, as should be expected for a correctly-trained model. The ``rough'' edges of this uncertainty are from data points that, while having a very similar freestream wind-speed, may have significantly different wind-directions, which also contribute to the predictive uncertainty. Secondly, while the majority of the influence on output power appears to be associated with the freestream wind-speed (as expected), the models have allocated some weight to the wind-directional features, resulting in the variance in the mean predictions for a given freestream wind-speed; this is mostly prevalent in the transitional region between minimum and maximum power, and matches intuition since wind-turbine wakes affect this region the most. However, in this region it is clear that significant deviation between model mean predictions and the test data remains. This deviation is likely to be the result of a number of factors, including the oversimplification of the wake-effect in assuming it can be approximated based on the wind-direction. In reality, this fails to account for atmospheric conditions such as wind shear, turbulence intensity, thermal stratification and air-density variations, which are known to have an impact on wake propagation \cite{nielson_using_2020,barthelmie_evaluation_2010}. Furthermore, in calculating the input features, the wind-direction (yaw angle) is assumed to be the same for all turbines---this is not the case, as there is a distribution of localised wind-directions across the wind-farm (yaw angles), which will likely change the wake patterns. There are also additional sources of noise within the data, such as sensor calibration errors in the yaw and wind-speed measurements which are used in calculating the population-level features, as well as possible changes in the underlying behaviour of the turbines, perhaps because of wear or damage during the period of the training data used. Additionally, \textit{curtailed}, \textit{shut-down} or even \textit{power-boosted} turbines upstream of a given turbine will further generate noise in the wake patterns.

Despite these problems, comparison of the west and east-side models yields some evidence that a level of turbine-dependent wind-directionality has been captured. For example, for the west-side turbine, it is not expected to be influenced by wakes when wind approaches from the western direction. Conversely, the east-side turbine is not expected to be influenced by wakes when wind blows from the east. In each case, the models appear to predict higher levels of power when these conditions are true, and less power when they are not, as one would expect to be the case in reality. While not shown here, these characteristics can also be seen for other turbine models around the perimeter of the wind-farm, to similar or lesser degrees.

To explore this model behaviour further, the learned model parameter vectors which result in differences in predicted power may be inspected. In this case, each of the vectors $\boldsymbol{\eta}$ and $\boldsymbol{\zeta}$ consist of 18 parameters, with parameters 1-6 associated with Feature One, 7-12 Feature Two and 13-18 Feature Three. Figure \ref{fig:weightcorrelations} shows the value of selected parameters from the $\boldsymbol{\eta}$ vector plotted for each turbine within the wind-farm. The colour of the points denotes the values of the parameters as indicated by the colour bars. 

For Parameter Four (associated with Feature One), there does not appear to be an obvious spatial correlation. However, while their magnitude is smaller than Parameter Four, Parameters Ten and 13 (associated with Features Two and Three), shown in subfigures \ref{fig:weight_10} and \ref{fig:weight_13}, show clear spatial correlation across the wind-farm. For Parameter Ten, there is an approximate negative correlation from east to west, while there is an approximate south to north negative correlation for Parameter 13. While not shown here, similar observations can be made for some of the remaining parameters in the $\boldsymbol{\eta}$ and $\boldsymbol{\zeta}$ vectors. These spatial differences in parameters represent the wake-effect captured by the NP models, and motivated the development of the metamodel, which aims to remove the requirement of having data from each individual turbine, by utilising these spatial correlations in the individual model parameters. The use of a hierarchical model in this case also allows ``statistical strength'' to be shared between turbine-specific parameters, via the population-level parameters.

\begin{figure}[H]
  \begin{subfigure}{\textwidth}
    \centering
    \includegraphics[width=.7\textwidth]{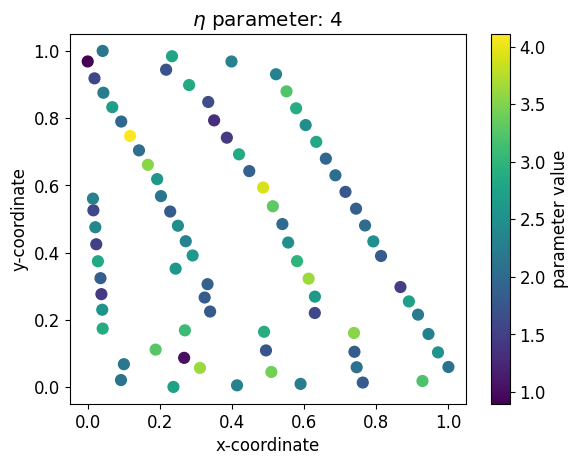}
    \caption{4th parameter in the learned $\boldsymbol{\eta}$ vectors.}
    \label{fig:weight_4}
  \end{subfigure}

\end{figure}
  \begin{figure}[H]\ContinuedFloat
  \begin{subfigure}{\textwidth}
    \centering
    \includegraphics[width=.7\textwidth]{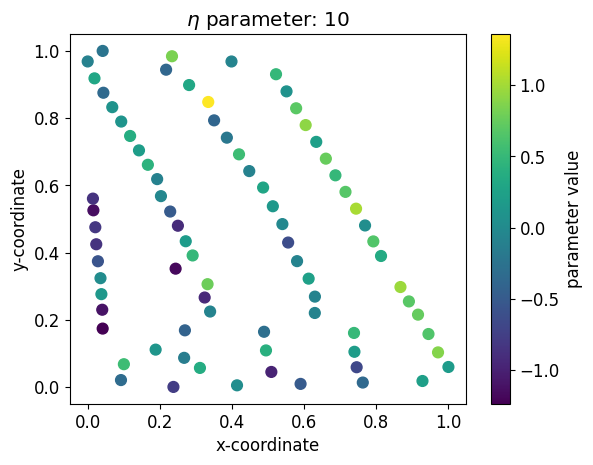}
    \caption{10th parameter in the learned $\boldsymbol{\eta}$ vectors.}
    \label{fig:weight_10}
  \end{subfigure}

\end{figure}
\begin{figure}[H]\ContinuedFloat
  \begin{subfigure}{\textwidth}
    \centering
    \includegraphics[width=0.7\textwidth]{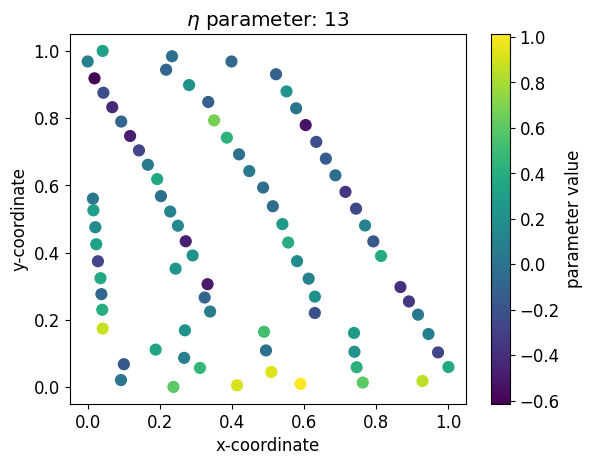}
    \caption{13th parameter in the learned $\boldsymbol{\eta}$ vectors.}
    \label{fig:weight_13}
  \end{subfigure}
  \caption{Mean parameter values of selected entries of the learned BSBR model $\boldsymbol{\eta}$ vectors, plotted by turbine location. Colours indicate the value of the parameters.}
  \label{fig:weightcorrelations}
  \end{figure}

\subsection{Metamodel}
The metamodel learns linear models over the parameter vectors, for turbine $x$ and $y$ coordinates. The right-hand plots in sub-Figures \ref{fig:weight4comparison} and \ref{fig:weight10comparison} show metamodel mean predicted values by colour, for $\boldsymbol{\eta}$ Parameters Four and Ten respectively, spatially. The black filled circles denote the locations of the turbines used for training the metamodel, while the unfilled black circles represent the locations of the remaining (unobserved) turbines that were also used in testing. These predictions are compared directly with the NP BSBR models, shown in the left-hand plots. As with Figure \ref{fig:weight_4}, there is no clear observable spatial correlation for Parameter Four seen for the NP model; there are also some potential outliers---such as those shaded more yellow in the upper half of the wind-farm. For the same parameter in the metamodel, the regularising effect of the linear model assumption removes the outliers in prediction, while perhaps only a weak correlation in learned values across the farm is observed. For Parameter Eight it can be seen that the metamodel has approximated the more clear observable correlation seen in the NP model parameters. These observations provide confirmation that the metamodel is working as intended.

\begin{figure}[H]
  \centering
  \begin{subfigure}{\textwidth}
    \centering
    \includegraphics[width=\textwidth]{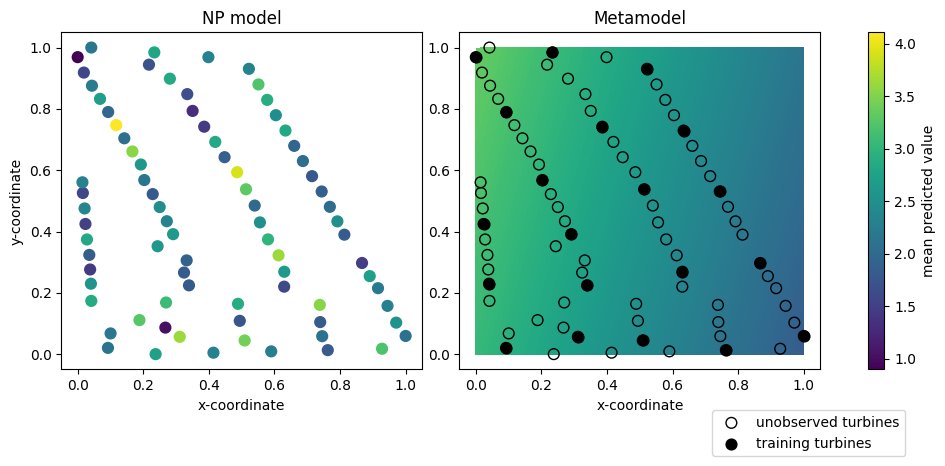}
    \caption{4th parameter in the learned $\boldsymbol{\eta}$ vectors.}
    \label{fig:weight4comparison}
  \end{subfigure}
\end{figure}
\begin{figure}[H]\ContinuedFloat
  \begin{subfigure}{\textwidth}
    \centering
    \includegraphics[width=\textwidth]{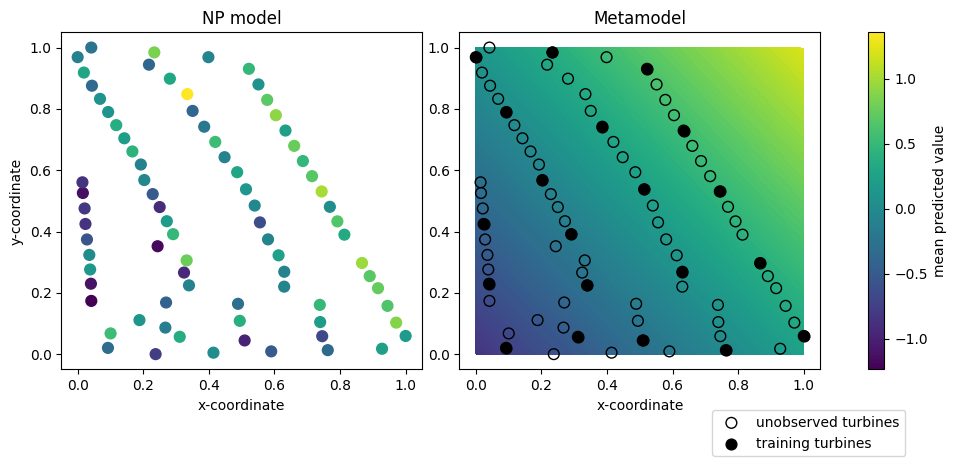}
    \caption{10th parameter in the learned $\boldsymbol{\eta}$ vectors.}
    \label{fig:weight10comparison}
  \end{subfigure}
  \caption{Mean predicted values for the Fourth (subplot (a)) and Tenth (subplot (b)) entries in the $\boldsymbol{\eta}$ parameter vector, plotted spatially for each turbine in wind-farm \textit{Ciabatta}. In each subplot, the left-hand plots show the mean predictions from the NP model, while the right-hand plots show the mean metamodel predictions. Filled circles represent turbines used in training, while unfilled circles represent turbines used in testing only. The points are shaded by their values.}
  \label{fig:metacorrelations}
\end{figure}

\subsubsection{Model Scores}

\begin{figure}[h]
  \begin{subfigure}{\textwidth}
  \includegraphics[width=\textwidth]{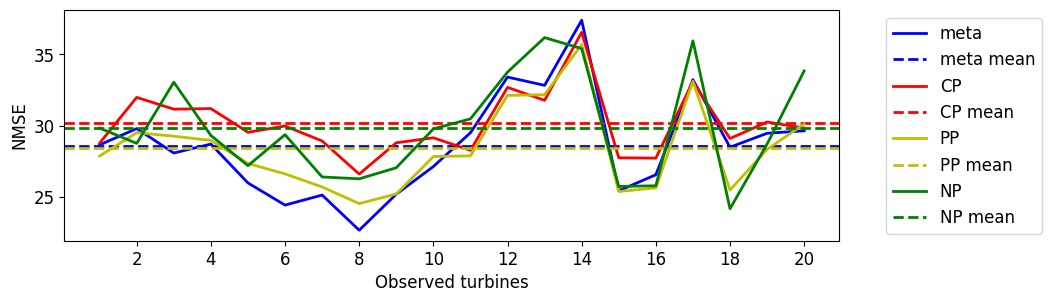}
  \caption{NMSE scores on the observed turbines}
  \label{fig:NMSE_observed}
  \end{subfigure}
  
  \bigskip
  \begin{subfigure}{\textwidth}
  \includegraphics[width=\textwidth]{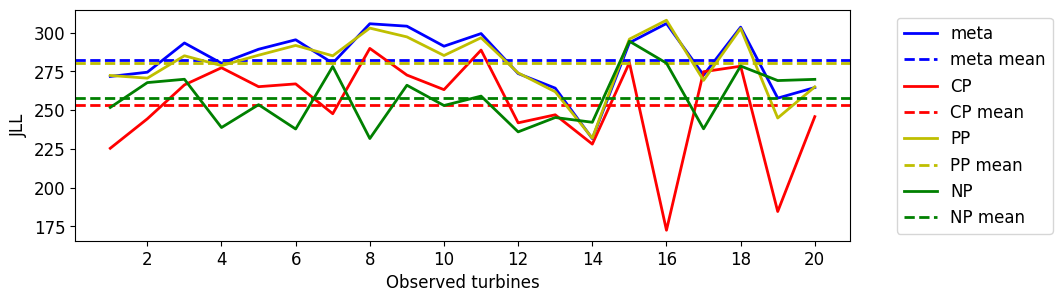}
  \caption{JLL scores on the observed turbines}
  \label{fig:JLL_observed}
  \end{subfigure}

  \caption{Comparison of model scores for each of the observed turbines in wind-farm \textit{Ciabatta}. Dashed horizontal lines indicate mean scores per model.}
  \label{fig:observed_scores}
\end{figure}

The metamodel is compared against the benchmark NP, PP and CP models, using the scoring metrics. In the case of the NMSE, a lower score indicates improved model performance, while for the JLL a higher score indicates an improved model fit. Firstly, Figure \ref{fig:NMSE_observed} shows the NMSE per turbine, for each model, for the \textit{observed} turbines only (i.e.\ those used in training). Similarly, Figure \ref{fig:JLL_observed} shows the JLL per turbine, for each model. For both metrics, it can be seen that on average, the metamodel and PP models score similarly to each other, and marginally better than both the CP and NP models, which themselves score similarly to each another. 

\begin{figure}[h]

  \begin{subfigure}{\textwidth}
  \includegraphics[width=\textwidth]{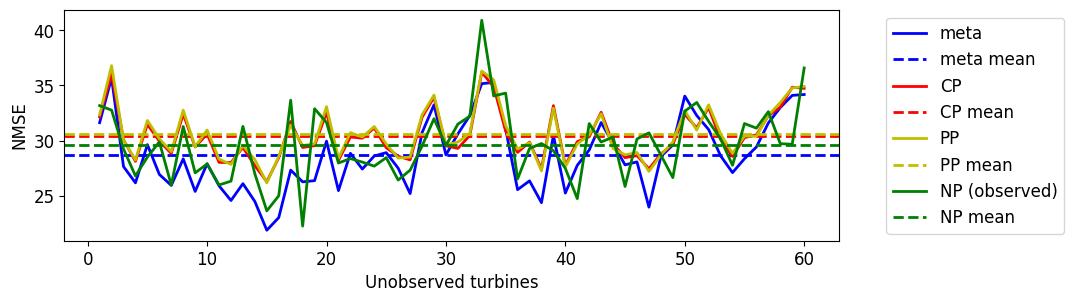}
  \caption{NMSE scores on the unobserved turbines}
  \label{fig:NMSE_unobserved}
  \end{subfigure}
  
  \bigskip
  
  \begin{subfigure}{\textwidth}
  \includegraphics[width=\textwidth]{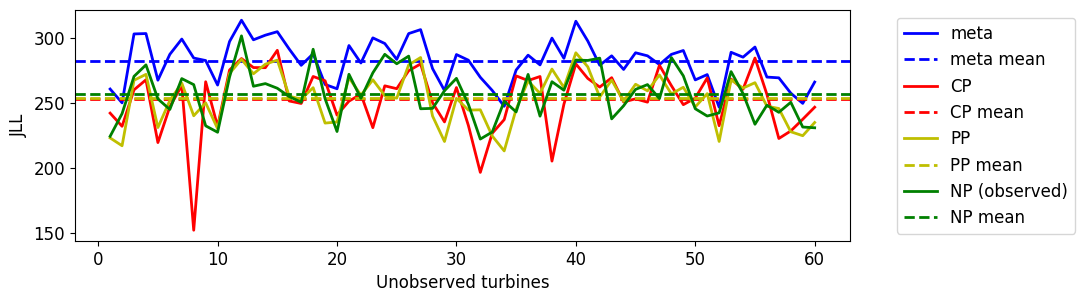}
  \caption{JLL scores on the unobserved turbines}
  \label{fig:JLL_unobserved}
  \end{subfigure}

  \caption{Comparison of model scores for each of the unobserved turbines in wind-farm \textit{Ciabatta}. Dashed horizontal lines indicate mean scores per model.}
  \label{fig:unobserved_scores}
\end{figure}

Secondly, Figures \ref{fig:NMSE_unobserved} and \ref{fig:JLL_unobserved} show the NMSE and JLL per turbine, for each model, for the \textit{unobserved} turbines; as a reminder, the exception to this is the NP models, which require data from every turbine, and so all turbines are individually observed in this case. For these otherwise unobserved turbines, it can be seen that on average the metamodel performs marginally better than all three of the other models for both the NMSE and JLL metrics, including the NP models despite not having observed data from these turbines in training. 

Across all observed and unobserved turbines, the CP model scores particularly poorly on the JLL metric in five instances: the observed turbines 16 and 19, and the unobserved turbines eight, 32 and 38. Since the CP model learns a single set of shared population-level parameters which are used for prediction, it is not able to adjust for turbine-specific differences, which may be the case in these instances. For the unobserved turbines, the PP model is limited in the same way as the CP model, since the shared population-level parameters must be used for making predictions; this explains their similar average scores. However unlike the CP model, the PP model does not share the same sharp drops in the JLL metric (turbines eight, 32 and 38), suggesting that partial-pooling of information provides improved adaptation to the model. In the case of the observed turbines, the PP model is able to use the learned turbine-specific weights in prediction, and so can make adjustments which improve the model scores relative to the CP model, which are of a similar performance to the metamodel. 


The metamodel combines strengths from all three approaches: information is pooled reducing the effective data-scarcity (like the CP and PP models), whilst turbine-specific adjustments may also be made (like the PP and NP models). Importantly however, the metamodel can additionally make adjustments for unobserved tasks, resulting in a distinct improvement in model performance for these tasks. Compared to the next best performing model for the unobserved turbines (the NP models), the metamodel scores were improved by 3.2\% and 9.8\% for NMSE and JLL metrics respectively.  



\begin{figure}[h]
  \centering
    \centering
    \includegraphics[width=\linewidth]{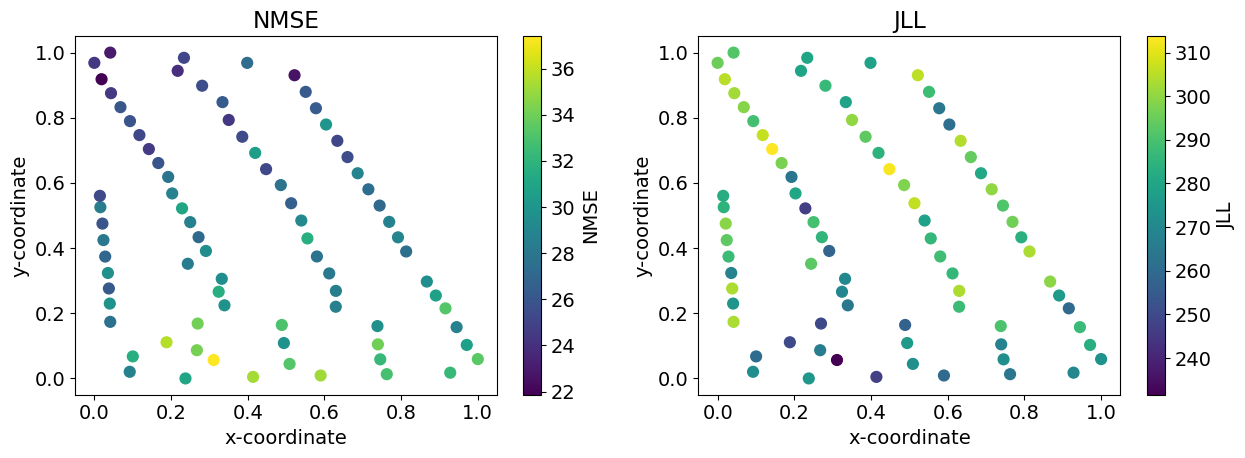} 
    \caption{Maps of the metamodel scores across the wind-farm. The left-hand plot shows the NMSE per turbine, while the right-hand plot shows the JLL per turbine.}
    \label{fig:score_maps}
\end{figure}

A final observation is that there are some relatively-high NMSE scores (and reduced JLL scores) across all models for the observed turbines 12--14, and the unobserved turbines 32--35, compared to the surrounding turbines; to explore this further, the scoring metrics for the metamodel are plotted spatially in Figure \ref{fig:score_maps}. Here, in general the scores appear to be worse towards the lower end of the wind-farm; since these turbines follow a less rigid spatial pattern than the turbines at the upper end, they are likely to be a source of additional noise in the spatial wake patterns of these turbines, which may make accurate prediction of output power using just the freestream wind speed and direction features used in this work more challenging.

\subsubsection{Predicting \textit{Unobserved} Turbines}
\begin{figure}[htbp]
  \centering
  \begin{subfigure}{.45\textwidth}
    \centering
    \includegraphics[width=\linewidth]{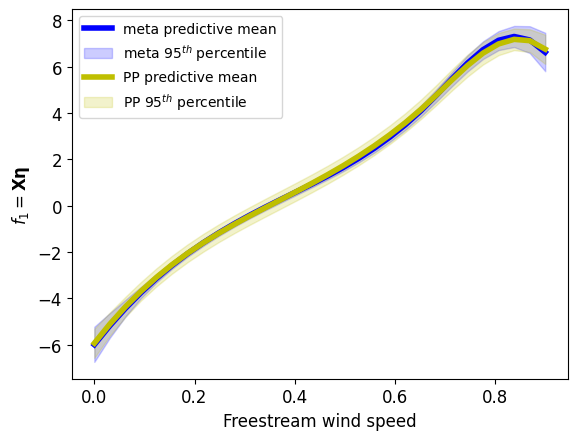} 
    \caption{$f_1$: metamodel and PP model}
    \label{fig:f1_a}
  \end{subfigure}
  \begin{subfigure}{.45\textwidth}
    \centering
    \includegraphics[width=\linewidth]{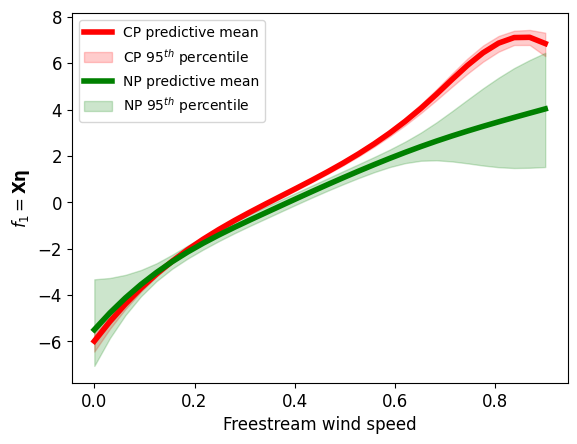} 
    \caption{$f_1$: CP and NP model}
    \label{fig:f1_b}
  \end{subfigure}
  \begin{subfigure}{.45\textwidth}
    \centering
    \includegraphics[width=\linewidth]{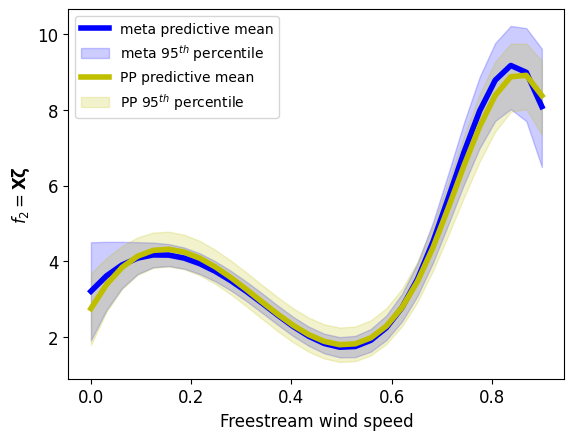} 
    \caption{$f_2$: metamodel and PP model}
    \label{fig:f2_a}
  \end{subfigure}
  \begin{subfigure}{.45\textwidth}
    \centering
    \includegraphics[width=\linewidth]{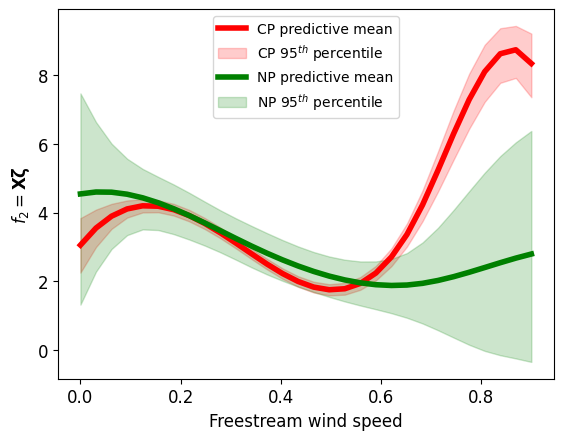} 
    \caption{$f_2$: CP and NP model}
    \label{fig:f2_b}
  \end{subfigure}
  \caption{Posterior predictive distributions for functions $f_1$ and $f_2$, for the \textit{unobserved} turbine on the east-side of the farm, using a gridded input over all three features. Averages were taken over Features Two and Three. Solid lines represent the mean prediction, while shaded areas show the $95^{th}$-percentile.}
  \label{fig:f_functions}
\end{figure}

The models are now compared in prediction, for an unobserved turbine---for consistency, the same turbine as previously predicted for in Figure \ref{fig:east_side} was chosen. Figures \ref{fig:f1_a}--\ref{fig:f2_b} first compare the learned linear models $f_1$ and $f_2$ across all four models. A gridded input over all three features was used, with averages of the predictive mean and $95^{th}$ percentile taken over the wind-directional features (Features Two and Three). Mean predictions are shown as solid lines, while $95^{th}$-percentiles are shown as shaded areas.

To aid interpretation, $f_1$ is proportional to the mean prediction as per Equation (\ref{eq:mu}), while $f_2$ is proportional to the precision (see Equation (\ref{eq:phi})), which is interpreted as the inverse of the variance. While the CP, PP and metamodels are broadly similar in these plots, there are significant differences in the function shape and $95^{th}$ percentile that can be seen for the NP model. As a result of the stratified sampling carried out on the training and test data, on a turbine-by-turbine basis there is relative data-scarcity at high freestream wind speeds; for an individual NP model such as this one, this scarcity is likely to result in both increased uncertainty in the trained model weights in this region of the curve, and a tendency for these weights to also be more biased towards their \textit{prior} distribution, which in this case were centred on zero. These factors likely both contribute towards $f_1$ being pulled downwards at higher wind speeds (and with increased uncertainty), and $f_2$ (proportional to precision) being relatively smaller (resulting in increased variance in the beta distribution). However, for both $f_1$ and $f_2$ the CP, PP and metamodels are not completely free of the issues of data-scarcity, as at high freestream wind speeds (beyond 0.8), the curves begin to decrease, which does not match the shape of the power-curve training data---power does not decrease at high wind speeds, nor does its variance increase after rated power is reached (assuming the cut-out wind speed is not reached). This model behaviour is perhaps a limitation of its current design, which could be improved by placing tighter, more positive priors for the higher freestream wind speed weights, which could be reasonably justified based on known behaviour of wind-turbines. More generally for the $f_1$ and $f_2$ functions, it can be said that the metamodel, PP and CP models benefit from pooling of information, resulting in narrower $95^{th}$ percentile predictions for both functions, despite not having observed data from this turbine. 

\begin{figure}[htbp]
  \centering
  \begin{subfigure}{.45\textwidth}
    \centering
        \includegraphics[width=\linewidth]{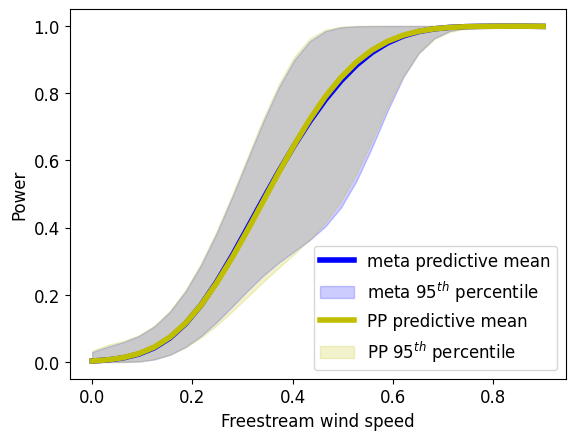} 
        \caption{Metamodel and PP model.}
        \label{fig:preds_a}
  \end{subfigure}
  \begin{subfigure}{.45\textwidth}
    \centering
        \includegraphics[width=\linewidth]{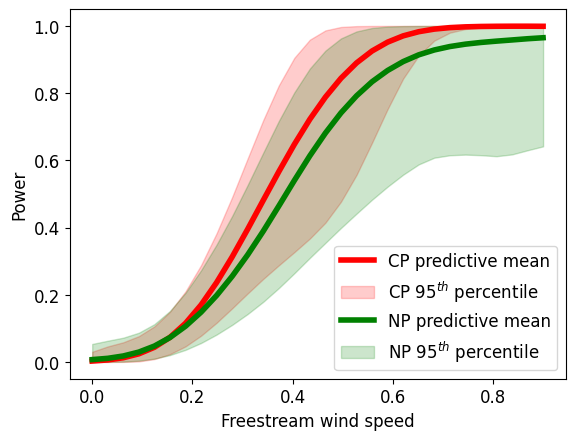} 
        \caption{CP and NP model.}
        \label{fig:preds_b}
  \end{subfigure}

  \caption{Probabilistic power-predictions for the \textit{unobserved} turbine, over gridded input data, averaged for the wind-direction features}
  \label{fig:unseen_preds}
\end{figure}

These observations can directly be seen in the full probabilistic power-predictions---shown in Figure \ref{fig:unseen_preds}. Here, the mean prediction of the NP model can be seen to be lower than the other models, with increased uncertainty at higher freestream wind speeds as indicated by the $f_1$ and $f_2$ functions. The CP, PP and metamodels all show broadly similar predictions, which are well-matched to the shape of power-curve data. In this space, the effects of the reducing $f_1$ and $f_2$ functions at high freestream wind speeds appears to be negligible.

\begin{figure}[h]
  \centering
    \centering
    \includegraphics[width=0.7\linewidth]{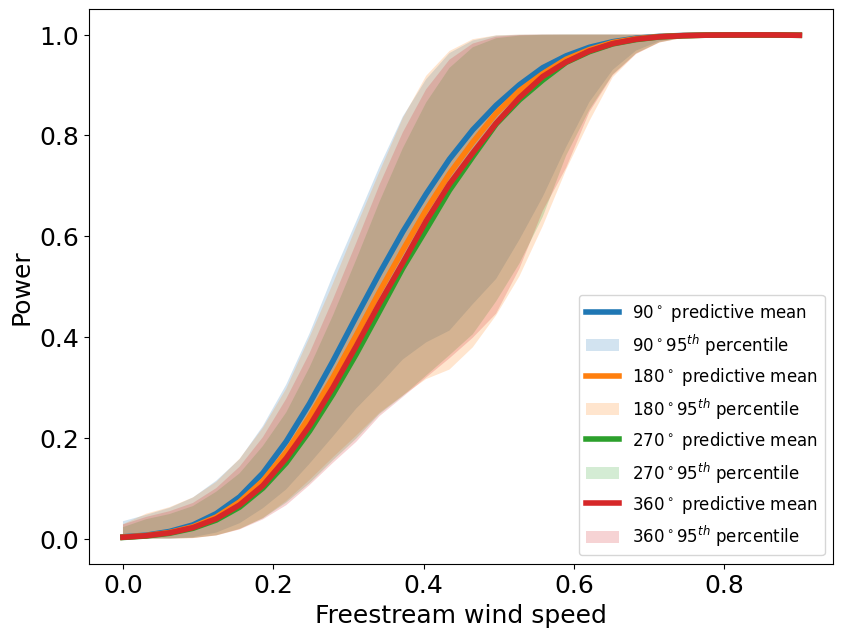} 
    \caption{Probabilistic power-predictions for the metamodel for the \textit{unobserved} turbine 35, over a gridded input of freestream wind-speeds, for easterly ($90^\circ$), southerly ($180^\circ$), westerly ($270^\circ$) and northerly ($360^\circ$) wind-directions.}
    \label{fig:unseen_meta_preds}
\end{figure}

Finally, to demonstrate how the metamodel is able to make location-specific adaptations in prediction for different wind-directions (which the CP and PP models cannot), Figure \ref{fig:unseen_meta_preds} shows power predictions for the same unobserved turbine, for easterly ($90^\circ$), southerly ($180^\circ$), westerly ($270^\circ$) and northerly ($360^\circ$) winds. As was shown for the model predictions in Figure \ref{fig:east_side}, the model mean prediction is greater in the wake-effected region of the power-curve during easterly winds (as compared to westerly winds), which matches intuition.

\subsection{Model Computational Complexity}\label{section:model_complexity}
To compare the computational complexity of the models, Table \ref{table:model_complexity} compares both the computational complexity of the likelihoods, and the number of learned parameters in each case.

\begin{table}[H]
\centering
\resizebox{\columnwidth}{!}{%
\begin{tabular}{|l|l|l|l|l|l|}
\hline
\textbf{Model}            & $T$ & $B$ & $N$ & \textbf{\begin{tabular}[c]{@{}l@{}}Likelihood \\ complexity\end{tabular}} & \textbf{\begin{tabular}[c]{@{}l@{}}Number of \\ parameters\end{tabular}} \\ \hline
No-pooling       & 80         & 18         & 250        & $\mathcal{O}(NT) = \mathcal{O}(20000)$                                                             & $2B \times T = 36 \times 80 = 2880$                                                                \\ \hline
Complete-pooling  & 20         & 18         & 250        & $\mathcal{O}(NT) = \mathcal{O}(5000)$                                                              & $2B = 36$                                                                 \\ \hline
Partial-pooling & 20         & 18         & 250        & $\mathcal{O}(NT) = \mathcal{O}(5000)$                                                              & $2BT + 2B + 2 = 758$                                                       \\ \hline
Meta model       & 20         & 18         & 250        & $\mathcal{O}(NT) = \mathcal{O}(5000)$                                                              & $8B = 144$                                                                 \\ \hline
\end{tabular}%
}
\caption{Comparison of the model likelihood complexity, and the number of parameters for each model. $T$: number of training turbines, $B$: width of the design matrix, $N$: data points per training turbine.}
\label{table:model_complexity}
\end{table}

While the number of iterations, and the model architecture influence overall computational complexity, the likelihood calculation is repeated every HMC iteration, and so should contribute significantly to the overall computational cost; in this case, since the CP, PP and metamodels are trained on 20 of the turbines, their likelihood complexity is one quarter of the overall cost of the 80 NP models. Additionally, the total number of parameters learned in each model varied significantly; the NP and PP models learn parameters for every turbine used in training (totalling 2880 and 758 respectively), while the CP and metamodels only learn parameters at the population-level (36 and 144 respectively). A higher number of parameters are also likely to increase the per-iteration cost, while also increasing the complexity of the \textit{posterior} geometry, which affects how efficiently HMC can move \cite{gelman2013bayesian}. In practice, model training varied in the order of hours for the CP, PP and metamodels, while it was in the order of a day for all 80 of the NP models, which roughly correlates with the increased total likelihood complexity, and number of parameters for these models.


\section{Conclusions}\label{section:conclusions}

A probabilistic, multivariate wind-turbine power-prediction model was developed, utilising a beta regression model with B-spline basis-functions (named the BSBR model). This BSBR model was shown to be successful in capturing both the heteroscedastic and asymmetric properties of power-curve data. When training the BSBR model on each turbine within a wind-farm, correlations were observed in the learned model parameters with respect to turbine locations, which were deemed to be associated with the directional dependency of wind-turbine wakes.

These spatial correlations in parameters were then leveraged via a ``metamodel'', which imposed functional constraints over turbine-specific parameters, as part of a hierarchical Bayesian model. This model design is able to learn from all turbines (tasks) simultaneously, and via the metamodel parameters, can be used to make adaptive probabilistic predictions for turbines not necessarily included in the training data, using their spatial coordinates alone. Across an 80-turbine wind-farm, the metamodel was found to outperform three benchmark models that utilised no-pooling, partial-pooling and complete-pooling of information, both in terms of normalised mean squared error (NMSE) and joint log-likelihood (JLL) in the majority of cases, while having a relatively low number of model parameters. 

Since the models are trained on farm-level input features to predict turbine-level output power, they are tuned (to a degree) on the specific environmental conditions (including wake-effects) of the wind-farm used in this work, which are also related to the specific layout of the wind-farm, as well as the turbine-type. If one wishes to apply this model to another wind-farm of interest, the model parameters should therefore be re-learned on data directly from that wind-farm in order to obtain valid results.

While in practice, training data for all turbines in a wind-farm are often available for power-curve prediction, the results suggest that the metamodel design is an efficient modelling solution for situations where data may be limited, and correlations could be expected in a variable of interest among a population of structures. Potential further applications in this domain include (but are not limited to): structural load assessment for fatigue-life estimation, condition monitoring and fault detection, and design for wind-farm layouts. As another perspective, suppose one wishes to measure a new variable of interest in a population of structures; this modelling approach could justify instrumenting just a subset of this population. By leveraging shared information across the subset of the population, the model may estimate the variable of interest with sufficient precision for the remaining structures without sensors, reducing the overall cost of sensor deployment.

As highlighted in Section \ref{section:bsbr_results}, the complexity of wake-effects cannot be fully captured by the wind-direction alone, which likely limits the performance of all the models in predicting power in the region between the cut-in and the rated wind-speeds. Future work could consider incorporating additional explanatory variables, such as wind-shear, turbulence-intensity, thermal-stratification and air-density variations, which may improve predictive performance. For the turbulence-intensity, if the standard-deviation of the wind-speed was available in the SCADA data (which is sometimes the case), it may be estimated as the ratio of the standard-deviation to the mean wind-speed \cite{gocmen_estimation_2016}. The incorporation of weather forecast (or hindcast) data may also help in gaining information for the additional variables, which could be obtained from a provider such as ECMWF \cite{ecmwf_2024}. The additional uncertainty associated with forecasted data does need to be considered however, as well as how to combine the 10-minute SCADA data with the weather data, which is likely to be coarser. If turbine-specific data are used, then the incorporation of computational fluid dynamics-based wake modelling---or a surrogate model with lower computational cost---may offer further improvements. The \textit{PyWake} wind-farm simulation tool \cite{pywake_2023} may be a practical option for such surrogate modelling.

A number of simplifying assumptions were made in relation to the training features, to help capture wake-effects from the data; these features may be improved if meteorological mast data were available, which would likely be a more accurate reflection of freestream wind conditions. Given the complexity of turbine wake physics, future work could also consider using physics-based inputs, to develop a hybrid data and physics-based approach. Datasets labelled with turbine operating regimes, and instances of damage or repair that may indicate a change in underlying turbine behaviour, may also help with data preprocessing to reduce possible sources of noise. Model design may also be improved by optimising the number of B-spline knots per feature, or their locations---perhaps by increasing the density of knots in the regions of the cut-in wind-speed and where rated power is reached, since these regions have increased curvature. Alternative spline-modelling approaches could also be considered. The use of nonlinear metamodels may also be worth considering, particularly if parameter spatial correlations are complex.

If one wishes to scale the model for larger wind-farms, or perhaps with more data, it may be practical to reduce the computational cost. One approach could be the use of variational inference (instead of HMC), where the \textit{posterior} is approximated as one of a family of known distributions \cite{gelman2013bayesian}; this however may come at the cost of accuracy. Keeping the total number of model parameters as low as reasonably practicable is also likely to reduce the computational burden.

\section{Data availability}\label{section:data_availability}
For the purpose of reproducibility, the \texttt{stan} code used in this work has been made available online at: \url{https://github.com/smbrealy/wind_mtl}. The data used are part of a Non Disclosure Agreement and cannot be shared.

\section{Acknowledgements}
The authors gratefully acknowledge the support of the UK Engineering and Physical Sciences Research Council (EPSRC) and the Natural Environment Research Council (NERC), via grant references EP/W005816/1, EP/R003645/1 and EP/S023763/1. The authors also gratefully acknowledge Vattenfall R\&D for their time and for providing access to the data used in this study. For the purpose of open access, the author(s) has/have applied a Creative Commons Attribution (CC BY) licence to any Author Accepted Manuscript version arising.



\bibliographystyle{elsarticle-num-names} 
\bibliography{cas-refs}





\end{document}